# Multi-view Face Analysis Based on Gabor Features[*]

Hongli Liu， Weifeng Liu[*]， Yanjiang Wang

*College of Information and Control Engineering in China University of Petroleum, Qingdao 266580, China*

**Abstract**

Facial analysis has attracted much attention in the technology for human-machine interface. Different methods of classification based on sparse representation and Gabor kernels have been widely applied in the fields of facial analysis. However, most of these methods treat face from a whole view standpoint. In terms of the importance of different facial views, in this paper, we present multi-view face analysis based on sparse representation and Gabor wavelet coefficients. To evaluate the performance, we conduct face analysis experiments including face recognition (FR) and face expression recognition (FER) on JAFFE database. Experiments are conducted from two parts: (1) Face images are divided into three facial parts which are forehead, eye and mouth. (2) Face images are divided into 8 parts by the orientation of Gabor kernels. Experimental results demonstrate that the proposed methods can significantly boost the performance and perform better than the other methods.

## 1   Introduction

Face recognition (FR) and face expression recognition (FER) have achieved attractive progress in the technology for human-machine interface, computer vision and image analysis. Methods of classification based on sparse representation [1] have been widely employed in facial analysis, such as KSVD [2], D-KSVD [3, 19], compressive sensing [4] and so on.

In the process of facial analysis, facial features extraction plays a pivotal role [5]. In the paper, we use Gabor wavelet coefficients of facial feature points as features. Gabor kernels can imitate human visual system and out-perform computational characteristics of other algorithms, which are similar to reflecting region of the human brain cortex simple cells [6]. Consequently, Gabor kernels have been widely applied in the fields of facial analysis. In [7] Lyons M used a multi-orientation multi-resolution set of Gabor filters to extract formation and construct a facial expression classifier. In [8] Chengjun Liu presented an independent Gabor features (IGFs) method and its application to face recognition. In [9] Lyons M J automatically classified facial images by using elastic graphs labelled with 2D Gabor wavelet features. In [10] Fasel B introduced the most prominent automatic facial expression analysis methods and systems presented in the literature. However these methods regard face or Gabor features as a whole view, which ignore the different contribution to FR or FER of different facial or Gabor features' parts.

[*] Project supported by the Natural Science Foundation for Youths of Shandong Province, China (No.ZR2011FQ016), the National Natural Science Foundation of China (No.61301242, 61271407), and the Fundamental Research Funds for the Central Universities, China University of Petroleum (East China) (No.13CX02096A), (No.CX2013057)
[*] Corresponding author.
Email address: liuwf@upc.edu.cn (Weifeng Liu)

In this paper, we present a multi-view facial analysis method which combines sparse representation and Gabor features. We treat different facial components or orientations of Gabor kernels as the corresponding "views". Some comparison experiments are conducted on JAFFE database [11]. The experimental results showed the attractive performance of the proposed methods: Gabor+ multi-components facial analysis [12, 20] method (GmCFA method) and multi- orientations Gabor + facial analysis method (mOGFA method). Figure 1 shows the flow path of GmCFA method, in which face images are decomposed into three components (forehead, eye, mouth) by the distribution of the 122 points.

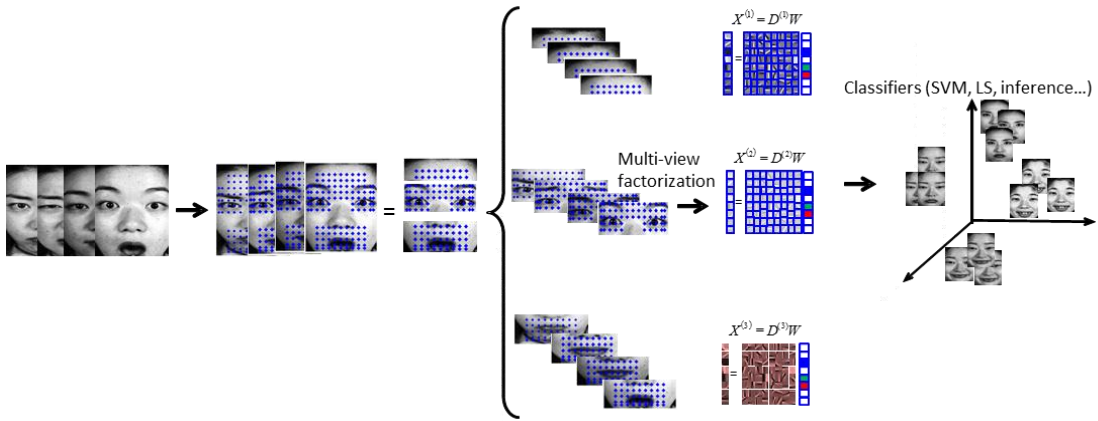

Figure 1. The framework of GmCFA

The remainder of this paper is organized as follows. Section 2 presents the role of Gabor features. Section 3 describes the framework of multi-view sparse coding. Section 4 follows the experiments and discussion. Finally in section 5, we conclude the paper and look forward the future work.

## 2   Gabor Features

Gabor kernels have been widely applied in the fields of facial features extraction by superior computational characteristics. 2-d Gabor wavelet [13] can be defined as:

$$\Psi_{u,v}(z) = \frac{\|k_{u,v}\|^2}{\sigma^2} e^{(\|k_{u,v}\|^2 \|z\|^2 / \sigma^2)} \left[ e^{ik_{u,v}z} \quad e^{\sigma^2/2} \right] \tag{1}$$

Where $k_{u,v} = k_v e^{i\phi_u}$, $k_v = \dfrac{k_{max}}{f^v}$ is frequency vector; $\phi_u = \dfrac{u\pi}{8}, \phi_u \in [0,\pi)$ is the switch orientation; $z = (x,y)$ is the coordinates of the image。 $v$ and $u$ represent the scale and orientation of Gabor kernels separately. In the paper, $v = \{0,1,2,3,4\}$, $u = \{0,1,2,\cdots,7\}$, so 40 Gabor coefficients were obtained on each point by convolution image with the Gabor kernels.

In the paper, 122 points (e.g. eye and mouth area) were employed as a fiducial facial mask. Fig. 2 shows the facial mask with distribution of the fiducial points.

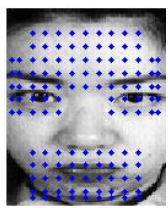

Figure 2. The fiducial facial mask

## 3 Multi-view Sparse Coding

Suppose Gabor features of $N$ face samples are divided into $P$ parts by facial organs or the orientations of Gabor kernels, where $x_i^p, p = 1,2,\cdots,P$ is the $p$ separated part of the $i$ face sample. We aim to find a sparse representation $W$ and the corresponding dictionary set $D$ by optimizing the following problem:

$$f = min_{D,W} \dfrac{1}{N}\sum_{p=1}^{P}\|X^{(p)} - D^{(p)}W\|_{Fro}^2 + \lambda\sum_{p=1}^{P}\|(D^{(p)})^T\|_{1,\infty} + \gamma\|W\|_{1,\infty} \qquad (2)$$

Where $X^{(p)} = [x_1^{(p)}, x_2^{(p)}, \cdots, x_N^{(p)}]$, $D^{(p)} = [D_1^{(p)}, D_2^{(p)}, \cdots, D_{N_d}^{(p)}]$ is the dictionary of the $p$ view. $\|D_j^{(v)}\|^2 \leqslant 1, 1 \leqslant j \leqslant N_d$, $N_d$ is number of dictionary atoms. $\lambda$ and $\gamma$ are parameters to balance the loss function and regularizations.

The optimization of problem (2) is convex w.r.t $D$ or $W$ separately, but not convex w.r.t $D$ and $W$ jointly. Consequently, we use alternating optimization [15] to figure out the problem. We solution the optimal problem by optimizing $W$ with $D$ fixed or optimizing $D$ with $W$ fixed. We present the optimization in detail as follows.

Given $D$ fixed, rewrite problem (2):

$$min_W \dfrac{1}{N}\sum_{p=1}^{P}\|X^{(p)} - D^{(p)}W\|_{Fro}^2 + \gamma\|W\|_{1,\infty} \qquad (3)$$

Given $W$ fixed, rewrite problem (2):

$$\min_D \frac{1}{N} \sum_{p=1}^{P} \|X^{(p)} - D^{(p)}W\|_{Fro}^2 + \lambda \sum_{p=1}^{P} \|(D^{(p)})^T\|_{1,\infty} \qquad (4)$$

Therefore we can find a sparse representation and the corresponding dictionary set by translate problem (2) into subproblem (3) and (4).

Algorithm 1 shows the steps of the problem (2) by alternating optimization.

step 1: Initialize $D, W$

step 2: repeat, until $|f - f_0| < tol$, $f$ is the optimization of the current iteration, $f_0$ is the optimization of the previous iteration.

step 3: Update $W$ by optimizing the problem (3).

step 4: Update $D$ by optimizing the problem (4).

## 3.1 Sparse representation $W$

Problem (3) can be rewritten as the following form

$$\min_W f(W) + g(W)$$

Where $f(W) = \frac{1}{N} \sum_{p=1}^{P} \|X^{(p)} - D^{(p)}W\|_{Fro}^2$, $g(W) = \gamma \|W\|_{1,\infty}$. $f(W)$ and $g(W)$ are both convex function. $f(W)$ is derivable, and $f(W)$ is continuous of Lipschitz. So subproblem (3) can be solved with a variant of Nesterov's first order method [16].

Algorithm 2 shows a convex optimization method of problem (3)

step 1: Initialize $W^0$, $\widetilde{W}^{(0)}$, 令 $\tau^{(0)} = 1$, $L_1 = \frac{1}{N} \sigma_{\max}(D^T D)$, $k = 0$;

step 2: repeat $k = 1, 2, \cdots$, until convergence;

step 3: Update $Z^{(k)}$, $Z^{(k)} \leftarrow \tau^{(k)} W^{(k)} + (1 - \tau^{(k)}) \widetilde{W}^{(k)}$;

step 4: Update $W^{(k+1)}$, $W^{(k+1)} \leftarrow \arg\min_W \|W - U^{(k)}\|_F^2 + \frac{\gamma}{\tau^{(k)} L_1} \|W\|_{1,\infty}$ \qquad (5)

where $U^{(k)} = W^{(k)} - \frac{1}{\tau^{(k)} L_1} (\frac{1}{N}(D^T D Z^{(k)} - D^T X))$;

step 5: Update $\widetilde{W}^{(k+1)}$, $\widetilde{W}^{(k+1)} = \tau^{(k)} W^{(k+1)} + (1 - \tau^{(k)}) \widetilde{W}^{(k)}$;

step 6: Update $\tau^{(k+1)} > 0$, $(\tau^{(k+1)})^{-2} - (\tau^{(k+1)})^{-1} = (\tau^{(k)})^{-2}$.

## 3.2 The multi-view dictionary $D$

Similarly, Problem (3) can be rewritten as the following form

$$\min_D f(D) + g(D)$$

Where $f(D) = \frac{1}{N} \sum_{p=1}^{P} \|X^{(p)} - D^{(p)}W\|_{Fro}^2$, $g(D) = \lambda \sum_{p=1}^{P} \|(D^{(p)})^T\|_{1,\infty}$. $f(D)$ and $g(D)$

are both convex function. $f(D)$ is derivable, and $f(D)$ is continuous of Lipschitz. So subproblem (4) can be solved by Algorithm 3.

Algorithm 3 shows a convex optimization method of problem (4)

step 1: Initialize $B^0$, $\tilde{B}^{(0)}$, 令 $\tau^{(0)} = 1$, $L_2 = \frac{1}{N}\sigma_{\max}(W^TW)$, $k = 0$;

step 2: repeat $k = 1,2,\cdots$, until convergence;

step 3: Update $Z^{(k)}$, $Z^{(k)} \leftarrow \tau^{(k)}B^{(k)} + (1-\tau^{(k)})\tilde{B}^{(k)}$;

step 4: Update $B^{(k+1)}$, $B^{(k+1)} \leftarrow \arg\min_B \|B - U^{(k)}\|_F^2 + \frac{\gamma}{\tau^{(k)}L_2}\|B\|_{1,\infty}$ (6)

where $U^{(k)} = B^{(k)} - \frac{1}{\tau^{(k)}L_2}(\frac{1}{N}(W^TWZ^{(k)} - W^TX))$;

step 5: Update $\tilde{B}^{(k+1)}$, $\tilde{B}^{(k+1)} = \tau^{(k)}B^{(k+1)} + (1-\tau^{(k)})\tilde{B}^{(k)}$;

step 6: Update $\tau^{(k+1)} > 0$, $(\tau^{(k+1)})^{-2} - (\tau^{(k+1)})^{-1} = (\tau^{(k)})^{-2}$.

Subproblem (5) and (6) can be efficiently solved using $l_1$ projection [17].

## 4 Experiments and Discussion

To evaluate the effectiveness of GmCFA method and mOGFA method, we apply linear SVM classifier and least squares (LS) to the integrated sparse codes obtained by the above algorithms. We conduct the experiments of FR and FER on JAFFE dataset. JAFFE dataset contains 213 facial images of 7 facial expressions from 10 Japanese females.

For each experiment, we select 70 facial images for testing, which contain each expression of each person, and the other facial images for training. We input the learned sparse codes into SVM [18] and LS classifiers to conduct face recognition and facial expression recognition. We compare the proposed multi-view facial analysis with single-view facial analysis which conducts experiments based on single facial component or single orientation of Gabor features. We also compare with the concatenation face analysis which considers the whole components or orientations as one part. For GmCFA method and mOGFA method, the parameters $\lambda, \gamma$ are tuned from set $\{0.001, 0.01, 0.1, 1\}$. And we use recognition rate as the criteria to evaluate the performance in our experiments.

In the paper, two methods are proposed to demonstrate the effectiveness of multi-view facial analysis. (1) GmCFA method, the normalized facial images are divided into three parts which are forehead, eye and mouth. (2) mOGFA method, the normalized face samples are divided into 8 views by the orientation of Gabor kernels.

## 4.1 GmCFA method

Figure 3 shows the average recognition rates with varying of the number atoms in FR and FER. Table 1 shows the comparison of expression recognition rates between single facial component and GmCFA method.

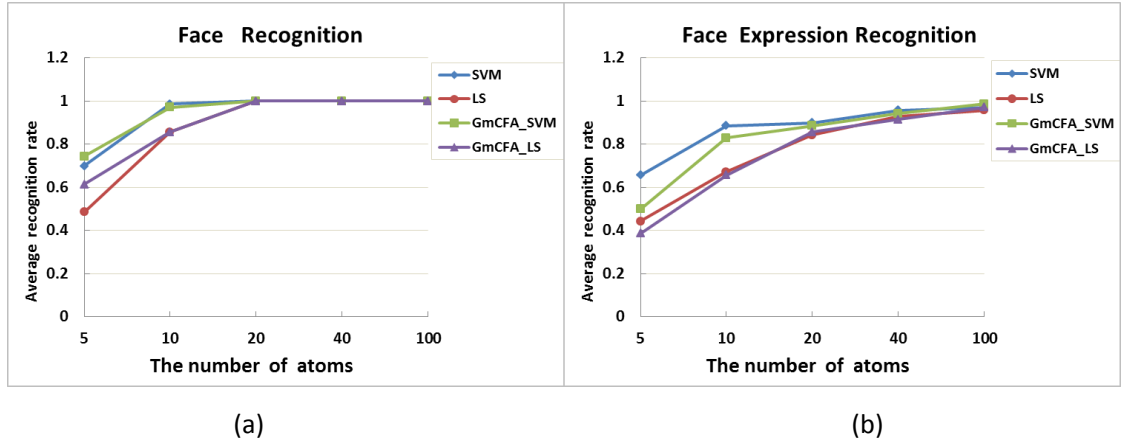

(a)  (b)

Figure 3. Average recognition rate with atom numbers

In Figure 3, SVM and LS represent the methods that the whole face images as one part. GmCFA_SVM, GmCFA_LS represent the methods that facial images are divided into three facial parts. In Figure 3(a), the average recognition rate of FR improves with the number of dictionary atoms increasing, and can reach up to 100% when the atoms are 20. In Figure 3(b), the average recognition rate of FER continues to rise with the number of dictionary atoms increasing. So we put an emphasis on FER of 100 atoms in the follow experiments.

Table 1. Expression recognition rates (%) of different methods

|  | AN | DI | FE | HA | NE | SA | SU | Aver |
|---|---|---|---|---|---|---|---|---|
| Forehead _LS | 100 | 80 | 60 | 100 | 90 | 70 | 100 | 85.71 |
| Eye_LS | 100 | 100 | 100 | 100 | 100 | 80 | 100 | 97.14 |
| Mouth_LS | 100 | 90 | 90 | 100 | 90 | 100 | 100 | 95.71 |
| GmCFA _LS | 100 | 100 | 90 | 100 | 100 | 90 | 100 | 97.14 |
| Forehead _SVM | 100 | 100 | 70 | 100 | 100 | 80 | 100 | 92.86 |
| Eye_SVM | 100 | 100 | 90 | 100 | 100 | 90 | 100 | 97.14 |
| Mouth_SVM | 100 | 90 | 100 | 90 | 90 | 100 | 100 | 95.71 |
| GmCFA _SVM | 100 | 100 | 90 | 100 | 100 | 100 | 100 | 98.57 |

In Table 1, AN, DI, FE, HA, NE, SA, SU denote the facial expression anger, disgust, fear, happy, neutral, sad, surprise respectively. The SVM and LS methods both work well in FER. GmCFA algorithm performs better than single facial component (forehead, eye, mouth) algorithm. And the average recognition rate of SVM algorithm outperforms LS algorithm in most cases.

## 4.2 mOGFA method

Table 2 shows the comparison of recognition results. Table 3 shows expression recognition results of different methods.

Table 2. Average Recognition rates (%) of different methods in FER

|     | Ori1  | Ori2  | Ori3  | Ori4  | Ori5  | Ori6  | Ori7  | Ori8  | Ori1-8 | mOGFA |
|-----|-------|-------|-------|-------|-------|-------|-------|-------|--------|-------|
| LS  | 90    | 95.71 | 92.86 | 91.43 | 91.43 | 95.7  | 95.71 | 90    | 95.71  | 97.14 |
| SVM | 92.86 | 92.86 | 91.43 | 92.86 | 92.86 | 92.86 | 97.14 | 94.29 | 95.71  | 97.14 |

In Table 2, Ori1, Ori2, Ori3, Ori4, Ori5, Ori6, Ori7, Ori8 denote 8 orientations of Gabor kernels. Ori1-8 denotes a whole matrix composed of 8 orientations. mOGFA denotes a multi-view matrix composed of 8 orientations. mOGFA method performs better than the other methods.

Table 3. Expression recognition rates (%) of different methods

|            | AN  | DI  | FE | HA  | NE  | SA  | SU  | Aver  |
|------------|-----|-----|----|-----|-----|-----|-----|-------|
| SRC+Gabor  | 100 | 90  | 80 | 90  | 100 | 90  | 100 | 92.9  |
| KSVD+Gabor | 100 | 100 | 80 | 90  | 100 | 90  | 90  | 92.9  |
| DKSVD+Gabor| 100 | 100 | 90 | 100 | 100 | 90  | 80  | 94.3  |
| GmCFA      | 100 | 100 | 90 | 100 | 100 | 100 | 100 | 98.57 |
| mOGFA      | 100 | 100 | 80 | 100 | 100 | 100 | 100 | 97.14 |

In Table 3, the average recognition rates of GmCFA method and mOGFA method are superior to the other methods [3] (SRC+Gabor, KSVD+Gabor, DKSVD+Gabor).

## 5 Conclusion

In this paper, we proposed GmCFA method and mOGFA method, which combine multi-view sparse representation and Gabor features for improving recognition results. Particularly, we conduct a batch of comparative experiments. Experimental results demonstrate that recognition results of our methods are obviously superior to other methods.